\documentclass[11pt,a4paper]{article}

\usepackage[utf8]{inputenc}
\usepackage[T1]{fontenc}
\usepackage{amsmath,amssymb}
\usepackage{booktabs}
\usepackage{graphicx}
\usepackage{hyperref}
\usepackage{xcolor}
\usepackage[margin=1in]{geometry}
\usepackage{natbib}
\usepackage{enumitem}
\usepackage{microtype}

\hypersetup{
    colorlinks=true,
    linkcolor=blue,
    citecolor=blue,
    urlcolor=blue
}

\title{Implicit Geographic Inference in LLM Medical Triage:\\
Language-Driven Disparities in Emergency Recommendations}

\author{
    Qi Han Wong\\
    \texttt{wongqihan@gmail.com}\\
    \url{https://github.com/wongqihan/ai-behavioral-experiments}
}

\date{June 2026}

\begin{document}

\maketitle

\begin{abstract}
We investigate whether large language models produce different medical triage recommendations for identical symptoms based solely on the language of the patient prompt. Using Gemini 3.5 Flash, we evaluate a neurological symptom profile (persistent headache, blurred vision, nausea) across six languages (English, Spanish, Chinese, Hindi, Japanese, Arabic) with 30 runs per condition ($n = 450$ total API calls). We find that the model recommends emergency room visits at rates ranging from 0\% (Japanese, Hindi) to 30\% (English, Arabic), despite assigning nearly identical severity scores (7.7--8.0/10) across all languages. Adding a single sentence specifying the patient's US location increases ER recommendations by up to 76.7 percentage points for non-English prompts, while the reverse anchor (English prompt with a Tokyo location) reduces the ER rate from 30\% to 6.7\%. A back-translation control (Japanese $\rightarrow$ English) produces ER rates comparable to the English baseline, confirming that the disparity is not caused by translation quality but by implicit geographic inference from the input language. We release the complete dataset, experiment code, and results.
\end{abstract}

\section{Introduction}

Large language models are increasingly deployed in healthcare-adjacent applications, including symptom checkers, triage assistants, and patient-facing chatbots \citep{singhal2023large, nori2023capabilities}. A critical assumption underlying these deployments is that the model's clinical recommendations are driven by the medical content of the input, not by superficial features such as the language in which the query is written.

We test this assumption directly. We present a single neurological symptom profile---persistent headache, blurred vision, morning nausea, and visual disturbances in a 38-year-old patient---to Gemini 3.5 Flash in six languages and measure the recommended clinical action (emergency room vs.\ doctor appointment). All prompts are semantically equivalent, manually authored to match native phrasing conventions in each language.

Our key finding is that the model uses \textbf{language as a proxy for geographic location}, then applies the healthcare norms of the inferred country:

\begin{itemize}[nosep]
    \item A Japanese prompt yields 0\% ER recommendations (consistent with Japan's clinic-first pathway)
    \item An English prompt yields 30\% ER recommendations (consistent with US defensive medicine norms)
    \item The model assigns severity scores of 7.7--8.0/10 across all languages, confirming equivalent clinical
    comprehension
\end{itemize}

We validate this mechanism through three controls: (1) adding a US location anchor to non-English prompts increases ER rates by 46.7--76.7 percentage points; (2) adding a Tokyo location anchor to the English prompt reduces the ER rate from 30\% to 6.7\%; and (3) back-translating the Japanese prompt to English produces ER rates comparable to the English baseline (36.7\% vs.\ 30\%), ruling out translation quality as a confound.

This behavior is consistent with the model learning country-specific healthcare practices from its training data. It produces appropriate recommendations for users located in the country associated with their language, but fails for immigrants, expatriates, tourists, and multilingual users whose language does not match their geographic location.

\textbf{Contributions:}
\begin{enumerate}[nosep]
    \item We demonstrate that a production LLM produces systematically different medical triage recommendations based on prompt language, independent of clinical content.
    \item We identify the mechanism as implicit geographic inference from language, validated through location anchoring and back-translation controls.
    \item We release the complete experiment code, prompts, and raw results for reproduction.\footnote{\url{https://github.com/wongqihan/ai-behavioral-experiments}}
\end{enumerate}

\section{Related Work}

\paragraph{Bias in Clinical NLP.}
Prior work has documented racial and gender biases in LLM-generated medical content. \citet{zack2024assessing} found that GPT-4 perpetuates race-based clinical decision-making when racial identifiers are included in prompts. \citet{omiye2023large} showed that large language models propagate race-based medicine by reproducing debunked claims about biological differences between racial groups. Our work extends this literature by demonstrating that \textit{language itself}---without any explicit demographic identifiers---functions as a proxy variable that shifts clinical recommendations.

\paragraph{Multilingual Disparities in LLMs.}
Cross-lingual performance gaps in LLMs are well-documented. Models consistently perform worse on non-English benchmarks \citep{ahuja2023mega, lai2023chatgpt}, and safety alignment degrades in low-resource languages \citep{deng2024multilingual, yong2024lowresource}. However, most prior work focuses on \textit{capability} gaps (the model understands less) rather than \textit{normative} shifts (the model applies different standards). Our finding is distinct: the model comprehends the symptoms equally well across languages (identical severity scores) but deliberately applies different action thresholds.

\paragraph{Geographic and Cultural Priors.}
LLMs encode geographic and cultural knowledge from their training corpora \citep{manvi2024large}. \citet{li2024culture} show that LLMs exhibit systematic cultural biases that correlate with the geographic distribution of their training data, producing culturally loaded outputs even when no cultural context is specified. Our work shows that this internalized geographic knowledge actively modifies clinical recommendations, creating a coupling between input language and inferred healthcare infrastructure.

\paragraph{AI Fairness in Healthcare.}
The deployment of AI in clinical settings raises fundamental fairness concerns \citep{chen2021ethical, obermeyer2019dissecting}. \citet{obermeyer2019dissecting} demonstrated that a widely-used commercial algorithm exhibited significant racial bias in predicting healthcare needs. Our finding adds a new dimension: language-based geographic inference as a source of disparate treatment in AI triage systems.

\section{Methodology}

\subsection{Model and Configuration}

We use Gemini 3.5 Flash (Google, May 2026 release) with temperature 0.3 and 30 runs per condition, totaling approximately 450 API calls. We chose a low temperature to maximize response consistency while preserving some stochastic variation for statistical analysis.

\subsection{Symptom Profile}

We use a single neurological symptom profile describing a 38-year-old patient with:
\begin{itemize}[nosep]
    \item Persistent headache for two weeks, unresponsive to painkillers
    \item Progressive blurred vision, especially when reading or using screens
    \item Morning nausea
    \item Occasional visual disturbances (spots)
    \item No prior history of similar symptoms
\end{itemize}

This profile was selected because it is clinically ambiguous---the symptoms are consistent with conditions ranging from tension headache (low urgency) to increased intracranial pressure (high urgency)---creating a decision boundary where contextual priors can influence the recommended action.

\subsection{Prompt Design}

All symptom prompts were manually authored (not machine-translated) to match native phrasing conventions in each language: English, Spanish, Chinese (Simplified), Hindi, Japanese, and Arabic. These six languages were selected to represent diverse writing systems, geographic regions, and healthcare infrastructure norms.

The system prompt enforces structured JSON output with the following fields:
\texttt{diagnosis} (English), \texttt{severity} (1--10), \texttt{urgency} (Emergency/\allowbreak{}Urgent/\allowbreak{}Routine), \texttt{action} (ER/\allowbreak{}Doctor appointment/\allowbreak{}Self-care), \texttt{tests} (list), and \texttt{reasoning} (English). The system prompt is always in English, regardless of the user prompt language.

\subsection{Test Conditions}

\begin{enumerate}[nosep]
    \item \textbf{Baseline:} Symptom prompts in 6 languages with no location context. $6 \times 30 = 180$ calls.
    \item \textbf{US Location Anchor:} Same prompts with an appended sentence in each language: ``Assume the patient is located in the United States.'' $6 \times 30 = 180$ calls.
    \item \textbf{Reverse Anchor:} English prompt with ``Assume the patient is located in Tokyo, Japan'' or ``Assume the patient is located in Mumbai, India.'' $2 \times 30 = 60$ calls.
    \item \textbf{Back-translation Control:} The Japanese prompt is translated to English by the same model, then triaged. This controls for translation quality. $1 \times 30 = 30$ calls.
\end{enumerate}

\subsection{Metrics}

Our primary outcome variable is the ER recommendation rate (proportion of runs where \texttt{action} = ``ER''). Secondary variables include average severity score and urgency classification. All 450 API calls returned valid, parseable JSON responses (0 parse failures). We report 95\% Wilson score confidence intervals for all proportions.

\section{Results}

\subsection{Baseline: Language Alone Shifts ER Rates}

Table~\ref{tab:baseline} shows ER recommendation rates across six languages with no location context. Severity scores are nearly identical (7.7--8.0), confirming that the model assesses the clinical danger equivalently regardless of language. The divergence is entirely in the recommended action.

\begin{table}[h]
\centering
\caption{Baseline ER recommendation rates by language (no location context, $n=30$ per language).}
\label{tab:baseline}
\begin{tabular}{lccccc}
\toprule
Language & ER Count & ER \% & 95\% CI & Avg Severity \\
\midrule
English  & 9/30  & 30.0\% & [16.7\%, 47.9\%] & 7.7 \\
Arabic   & 9/30  & 30.0\% & [16.7\%, 47.9\%] & 8.0 \\
Chinese  & 6/30  & 20.0\% & [9.5\%, 37.3\%]  & 8.0 \\
Spanish  & 4/30  & 13.3\% & [5.3\%, 29.7\%]  & 7.8 \\
Hindi    & 0/30  & 0.0\%  & [0.0\%, 11.4\%]  & 8.0 \\
Japanese & 0/30  & 0.0\%  & [0.0\%, 11.4\%]  & 8.0 \\
\bottomrule
\end{tabular}
\end{table}

The confidence intervals for English/Arabic (30\%) and Hindi/Japanese (0\%) do not overlap, confirming a statistically meaningful separation. The mid-range languages (Chinese at 20\%, Spanish at 13.3\%) have overlapping CIs with both groups and cannot be statistically distinguished from either at $n=30$.

\subsection{US Location Anchor Shifts Non-English Recommendations}

Adding a single sentence specifying a US location causes ER rates to surge across all non-English languages (Table~\ref{tab:anchor}). English shows minimal change (+10 pp), consistent with the model already assuming a US location for English-language queries.

\begin{table}[h]
\centering
\caption{Effect of US location anchor on ER recommendation rates ($n=30$ per condition).}
\label{tab:anchor}
\begin{tabular}{lcccc}
\toprule
Language & Default ER\% & + US Anchor ER\% & Shift & Anchor 95\% CI \\
\midrule
Chinese  & 20.0\% & 96.7\% & +76.7 pp & [83.3\%, 99.4\%] \\
Hindi    & 0.0\%  & 73.3\% & +73.3 pp & [55.6\%, 85.8\%] \\
Arabic   & 30.0\% & 90.0\% & +60.0 pp & [74.4\%, 96.5\%] \\
Spanish  & 13.3\% & 70.0\% & +56.7 pp & [52.1\%, 83.3\%] \\
Japanese & 0.0\%  & 46.7\% & +46.7 pp & [30.2\%, 63.9\%] \\
English  & 30.0\% & 40.0\% & +10.0 pp & [24.6\%, 57.7\%] \\
\bottomrule
\end{tabular}
\end{table}

The English shift (+10.0 pp) has heavily overlapping confidence intervals with the English baseline and is not statistically significant, while all non-English shifts produce non-overlapping intervals with their respective baselines.

\subsection{Reverse Anchor Confirms the Mechanism}

Applying a foreign location anchor to an English prompt reverses the effect (Table~\ref{tab:reverse}). English + Tokyo produces an ER rate of 6.7\%, and English + Mumbai produces 0.0\%, both substantially below the English baseline of 30.0\%.

\begin{table}[h]
\centering
\caption{Reverse anchor: English prompt with non-US location ($n=30$ per condition).}
\label{tab:reverse}
\begin{tabular}{lccc}
\toprule
Condition & ER Count & ER \% & 95\% CI \\
\midrule
English (baseline)            & 9/30 & 30.0\% & [16.7\%, 47.9\%] \\
English + Tokyo anchor        & 2/30 & 6.7\%  & [1.8\%, 21.3\%]  \\
English + Mumbai anchor       & 0/30 & 0.0\%  & [0.0\%, 11.4\%]  \\
\bottomrule
\end{tabular}
\end{table}

This result confirms that language alone does not drive the disparity. Explicitly overriding the inferred location changes the recommended action, even when the prompt language remains English.

\subsection{Back-Translation Rules Out Comprehension Gaps}

The Japanese prompt, translated to English by the same model and then triaged, produces an ER rate of 36.7\% (Table~\ref{tab:backtrans}). The confidence interval [21.9\%, 54.5\%] overlaps almost entirely with the English baseline [16.7\%, 47.9\%], confirming that the model comprehends the Japanese symptoms equivalently.

\begin{table}[h]
\centering
\caption{Back-translation control: Japanese prompt translated to English ($n=30$).}
\label{tab:backtrans}
\begin{tabular}{lccc}
\toprule
Condition & ER Count & ER \% & 95\% CI \\
\midrule
Japanese (original)                & 0/30  & 0.0\%  & [0.0\%, 11.4\%]  \\
Japanese $\rightarrow$ English     & 11/30 & 36.7\% & [21.9\%, 54.5\%] \\
\bottomrule
\end{tabular}
\end{table}

The 0\% ER rate for the Japanese prompt is not caused by the model failing to understand the symptoms. It is caused by the model inferring a Japanese location and applying Japanese healthcare routing norms.

\section{Discussion}

\subsection{Mechanism: Language $\rightarrow$ Location $\rightarrow$ Norms $\rightarrow$ Action}

Our results are consistent with the following inference chain:

\begin{center}
\textit{Input language} $\rightarrow$ \textit{Inferred country} $\rightarrow$ \textit{Regional healthcare norms} $\rightarrow$ \textit{Recommended action}
\end{center}

\begin{itemize}[nosep]
    \item \textbf{Japanese} $\rightarrow$ Japan $\rightarrow$ clinic-first pathway $\rightarrow$ ``Doctor appointment''
    \item \textbf{English} $\rightarrow$ USA $\rightarrow$ defensive medicine norms $\rightarrow$ ``ER'' (30\% of the time)
    \item \textbf{Hindi} $\rightarrow$ India $\rightarrow$ conservative triage $\rightarrow$ ``Doctor appointment''
\end{itemize}

Notably, the model does not mention geography or location in its reasoning field. The reasoning text across all languages describes the same clinical picture (intracranial pressure, papilledema risk, need for MRI). The geographic inference is implicit---it shapes the recommended action without surfacing in the model's stated rationale.

\subsection{Who Is Harmed?}

This behavior is appropriate for a monolingual user located in the country associated with their language. A Japanese speaker in Tokyo \textit{should} receive recommendations aligned with Japanese healthcare infrastructure. The failure mode arises for users whose language does not match their location:

\begin{itemize}[nosep]
    \item A Hindi-speaking immigrant in San Francisco receives 0\% ER recommendations for symptoms that would generate 30\% ER recommendations in English.
    \item An English-speaking expatriate in Tokyo receives 30\% ER recommendations despite being in a healthcare system where clinic-first is the standard pathway.
    \item Multilingual users receive different recommendations depending on which language they choose to describe their symptoms.
\end{itemize}

\subsection{Mitigation}

The most direct mitigation is to explicitly anchor the patient's geographic location in the system prompt. Our US anchor results demonstrate that the model already knows how to apply location-appropriate norms when given explicit context. The fix is not to remove geographic awareness from the model, but to decouple it from language:

\begin{quote}
\textit{``The patient is located in [country]. Use [country]'s healthcare system norms for your recommendation.''}
\end{quote}

Alternatively, when the patient's location is unknown, the model should surface the uncertainty rather than silently inferring it. A triage system could prompt: ``Where are you currently located?'' before generating a recommendation, analogous to how a human clinician in an unfamiliar setting would ask before advising.

\section{Limitations}

\begin{enumerate}[nosep]
    \item \textbf{Single model.} We tested only Gemini 3.5 Flash. The behavior may differ across model families (GPT-4, Claude, Llama). Cross-model replication is needed to establish generality.
    \item \textbf{Single scenario.} We tested one neurological symptom profile. The effect may not generalize to other clinical presentations (e.g., chest pain, psychiatric symptoms, pediatric cases). Different symptom profiles may have different degrees of geographic sensitivity.
    \item \textbf{Prompt equivalence.} While prompts were manually authored to match native phrasing, they were not independently verified by medical professionals or native speakers in each language. Subtle pragmatic differences in symptom description conventions could contribute to the observed disparities.
    \item \textbf{No human baseline.} We did not compare the model's recommendations to those of human clinicians from each country. Without a ground truth, we cannot determine which language's recommendations are ``correct''---only that they differ.
    \item \textbf{Sample size.} At $n=30$ per condition, some pairwise comparisons (e.g., English vs.\ Spanish baseline) have overlapping confidence intervals. Larger samples would be needed to establish statistical significance for mid-range comparisons.
    \item \textbf{English system prompt.} The system prompt was always in English, which may introduce an asymmetry for non-English user prompts. Testing with language-matched system prompts could reveal additional effects.
\end{enumerate}

\section{Conclusion}

We demonstrate that a production large language model produces systematically different medical triage recommendations based on prompt language, even when the clinical content is identical. The mechanism is implicit geographic inference: the model uses language as a proxy for the patient's location and applies region-specific healthcare norms. This behavior is validated through location anchoring experiments and a back-translation control.

For developers deploying LLMs in medical triage or healthcare-adjacent applications, our findings suggest a concrete mitigation: decouple geographic context from language by explicitly specifying the patient's location in the system prompt. Language should determine \textit{how} the model communicates, not \textit{what} it recommends.

Code, data, and full results are available at:\\
\url{https://github.com/wongqihan/ai-behavioral-experiments}

\section*{Acknowledgments}


\bibliographystyle{plainnat}

\end{document}